\pdfoutput=1

\documentclass[11pt]{article}

\usepackage[]{acl}
\usepackage{bbm}
\usepackage{times}
\usepackage{calc}
\usepackage{graphicx}
\usepackage{tabularx}
\usepackage{multirow}
\usepackage{booktabs}
\usepackage{latexsym}
\usepackage{amsmath}
\usepackage{algorithm}
\usepackage{algpseudocode}
\usepackage{bbm}

\usepackage[T1]{fontenc}
\usepackage{array}
\usepackage[utf8]{inputenc}
\usepackage{threeparttable} 

\usepackage{listings}
\usepackage{xcolor}
\usepackage{lipsum}
\definecolor{codegreen}{rgb}{0,0.6,0}
\definecolor{codegray}{rgb}{0.5,0.5,0.5}
\definecolor{codepurple}{rgb}{0.58,0,0.82}
\definecolor{backcolour}{rgb}{0.95,0.95,0.92}

\lstdefinestyle{mystyle}{
    backgroundcolor=\color{backcolour},   
    commentstyle=\color{codegreen},
    keywordstyle=\color{magenta},
    numberstyle=\tiny\color{codegray},
    numbers=none,
    stringstyle=\color{codepurple},
    basicstyle=\ttfamily\footnotesize,
    breakatwhitespace=false,         
    breaklines=false,                 
    captionpos=b,                    
    keepspaces=true,                 
    numbers=left,                    
    numbersep=5pt,                  
    showspaces=false,                
    showstringspaces=false,
    showtabs=false,                  
    tabsize=2
}
\usepackage{microtype}

\newcommand\model{\textsc{F-coref}}

\definecolor{darkspringgreen}{rgb}{0.09, 0.45, 0.27}

\title{\model{}: Fast, Accurate and Easy to Use Coreference Resolution}

\author{Shon Otmazgin\textsuperscript{1} \quad 
        Arie Cattan\textsuperscript{1} \quad
        Yoav Goldberg\textsuperscript{1,2} \quad \\ 
        \textsuperscript{1}Computer Science Department, Bar Ilan University \\ 
        \textsuperscript{2}Allen Institute for Artificial Intelligence \\ 
  {\normalsize\tt  \{shon711,arie.cattan,yoav.goldberg\}@gmail.com} \\
 }

\begin{document}
\maketitle

\begin{abstract}

We introduce \textit{fastcoref}, a python package for fast, accurate, and easy-to-use English coreference resolution. The package is pip-installable, and allows two modes: an accurate mode based on the \textsc{LingMess} architecture, providing state-of-the-art coreference accuracy, and a substantially faster model, \model{}, which is the focus of this work. \model{} allows to process 2.8K OntoNotes documents in 25 seconds on a V100 GPU (compared to 6 minutes for the \textsc{LingMess} model, and to 12 minutes of the popular AllenNLP coreference model) with only a modest drop in accuracy.
The fast speed is achieved through a combination of distillation of a compact model from the LingMess model, and an efficient batching implementation using a technique we call leftover batching.\footnote{\url{https://github.com/shon-otmazgin/fastcoref}} 

\end{abstract}
\section{Introduction}

Coreference Resolution consists of identifying textual mentions that refer to the same entity in a given text \citep{karttunen-1969-discourse-referents}. This fundamental NLP task can benefit various applications such as Information Extraction~\citep{luan-etal-2018-multi, li-etal-2020-gaia, jain-etal-2020-scirex}, Question Answering~\citep{dasigi-etal-2019-quoref, chen-durrett-2021-robust}, Machine Translation~\citep{stojanovski-fraser-2018-coreference, voita-etal-2018-context}, and Summarization~\citep{christensen-etal-2013-towards, falke-etal-2017-concept, pasunuru-etal-2021-efficiently}. However, compared to other core tasks such as POS Tagging, named-entity recognition or syntactic parsing, existing packages and state-of-the-art models for coreference resolution are challenging to apply: there are few easy-to-use packages implementing state-of-the-art models, and the available packages consume a lot of GPU memory, and take very long to process each document. For example, the coreference model in the popular AllenNLP package \cite{Gardner2017AllenNLP}, implementing the model of \citet{joshi-etal-2020-spanbert}, requires 27GB of GPU memory and takes 12 minutes to process the 2.8K documents of the OntoNotes corpus, on a V100 GPU. 

In this work, we introduce \model{}, a new open source Python package for simply running an efficient coreference model using a few lines of code. \model{} predicts coreference clusters 29 times faster than the AllenNLP model (processing the OntoNotes corpus in 25 seconds) and requires only 15\% of its GPU memory use, with only a small drop in performance (78.5 vs 79.6 average F1). The package also includes \textsc{LingMess} \cite{Otmazgin2022LingMessLI}, a state-of-the-art coreference model, which is almost twice as fast as the AllenNLP model, while being more accurate (81.4 average F1), under the same API.

To achieve \model{}'s speed, we use two additive techniques: model distillation of the strong-but-slow \textsc{LingMess} model using large unlabeled data, and an effective batching technique that reduces the number of padded tokens in a batch. 

\section{The \model{} API}

The \textit{fastcoref} Python package is pip installable (\texttt{pip install fastcoref}) and provides an easy and fast API for coreference information with only few lines of code without any prepossessing steps.

The \model{} constructor initializes our pretrained model on a single device:

\begin{lstlisting}[language=python,numbers=none]
from fastcoref import FCoref

model = FCoref(device='cuda:0')
\end{lstlisting}

\noindent The main functionally of the package is the \textit{predict} function, which accepts a list of texts.

\begin{lstlisting}[language=python,numbers=none]
preds = model.predict(
    texts=['We are so happy to see you 
    using our coref package.
    This package is very fast!']
)
\end{lstlisting}

\noindent The return value of the \textit{predict} function is a list of \textit{CorefResult} objects, from which one can extract the coreference clusters (either as strings or as character indices over the original texts), as well as the logits for each corefering entity pair:

\begin{lstlisting}[language=python,numbers=none]
preds[0].get_clusters(as_strings=False)
> [[(0, 2), (33, 36)], 
   [(33, 50), (52, 64)]
  ]

preds[0].get_clusters()
> [['We', 'our'], 
   ['our coref package', 'This package']
  ]
 
preds[0].get_logit(
    span_i=(33, 50), span_j=(52, 64)
)

> 18.852894
\end{lstlisting}

\noindent  Processing can be applied to a collection of texts of any length in a batched and parallel fashion:

\begin{lstlisting}[language=python,numbers=none]
texts = ['text 1', 'text 2',.., 'text n']

# control the batch size 
# with max_tokens_in_batch parameter

preds = model.predict(
    texts=texts, max_tokens_in_batch=100
)
\end{lstlisting}

The \texttt{max\_tokens\_in\_batch} parameter can be used to control the speed vs. memory consumption tradeoff, and can be tuned to maximize the utilization of the associated hardware.

To control speed vs. accuracy tradeoff, use the larger but more accurate \textsc{LingMess} model, simply import LingMessCoref instead of FCoref:

\begin{lstlisting}[language=python,numbers=none]
from fastcoref import LingMessCoref

model = LingMessCoref(device='cuda:0')
\end{lstlisting}

On top of the provided models, the package also provides a custom SpaCy component that can be plugged into a SpaCy(V3) pipeline~\citep{spacy}. Additionally, the package includes a \texttt{CorefTrainer} for training and distilling coreference model on your own data, opening the possibility for fast and accurate coreference models for additional languages and domains. 











To summarize, the package provides a simple API that makes predicting coreference entities straightforward and easy-to-use. The package supports any text length as input, and performs efficient batching. The package's \model{} model is 29 times faster and 4 times smaller than the popular coreference model in the AllenNLP package, while the provided \textsc{LingMess} mode is twice as fast the AllenNLP implementation, and more accurate.

\section{Background: Neural Coreference}
\label{sec:bg}

\citet{lee-etal-2017-end} present the first end-to-end model that jointly learns mention detection and coreference decision. Successive follow-up works kept improving performance through the incorporation of widely popular pretrained architectures~\citep{lee-etal-2018-higher, joshi-etal-2019-bert, kantor-globerson-2019-coreference, joshi-etal-2020-spanbert}. However, as the dimensionality of contextualized encoders increases, keeping in memory all possible span representations becomes highly costly and computationally untractable for long documents.

\subsection{Faster Neural Coreference}
Several methods have been proposed to address this memory constraint at the cost of the computation time and a slight performance deterioration~\citep{xia-etal-2020-incremental, toshniwal-etal-2020-learning, thirukovalluru-etal-2021-scaling}. The \emph{s2e} model of \citet{kirstain-etal-2021-coreference} managed to improve computation time with a slight \emph{increase} in accuracy.

\paragraph{s2e} Our \model{} is based on the architecture of the \emph{s2e} model by \citet{kirstain-etal-2021-coreference}. Like other neural coreference models, \emph{s2e} scores each pair of spans in the text to be co-referring to each other. However, in order to achieve lower memory footprint s2e moves to representing each span as a function of its start and end tokens. Consequently, the model avoids holding vector representation for each of the $O(n^2)$ spans in memory, and instead stores only $O(n)$ vectors. This reduced memory footprint allows it to handle longer sequences.

The \emph{s2e} architecture includes three components: (1) Longformer \cite{Beltagy2020LongformerTL}, a contextualized encoder; (2) a parameterized \emph{mention scoring function} $f_m$; and (3) a parameterized \emph{pairwise antecedent scoring function} $f_a$. To score any pair of spans to be co-referring, the model starts by encoding the text using Longformer into vectors $x_1, ..., x_n$. Using these vectors, for each possible span $q=(x_k, x_\ell)$ the mention scoring function $f_m(q)$, scores how likely $q$ (``query'') being a mention. Then for a pair of spans $c=(x_i, x_j)$, $q=(x_k, x_\ell)$ where $c$ (``candidate'') appears before $q$, the pairwise antecedent scoring function, $f_a(c,q)$, scores how likely is $c$ being an antecedent of $q$. In practice, to avoid complexity of $\mathcal{O}(n^4)$, the antecedent function scores only the $\lambda T$ spans with highest mention scores (where $T$ is the number of tokens). Finally, the final pairwise score for a coreference link between $c$ and $q$ is composed by the score of $q$ being a mention, $c$ being a mention, and how likely is $c$ being an antecedent of $q$:

\begin{align*}
F(c, q) = 
\begin{cases}
f_m(c) + f_m(q) + f_a(c, q) & c \neq \varepsilon \\
0 & c = \varepsilon 
\end{cases}
\end{align*}
where $\varepsilon$ is the null antecedent. 

The computation of $f_m$ and $f_q$ for the entire sequence can be efficiently batched.

\paragraph{Word-level coreference} \citet{dobrovolskii-2021-word} proposed moving from scoring pairs of spans to scoring pairs of words, establishing coreference relations between the words, and then expanding each of the relevant words into their mention boundaries. This reduces the model complexity from $\mathcal{O}(n^4)$ to $\mathcal{O}(n^2)$.

\subsection{What remains slow?}

While the \emph{s2e} and the word-level models are considered lightweight and efficient, and substantally improve in speed over \citet{joshi-etal-2019-bert}, their computation time is still dominated by their expensive contextualized encoding stage. They also use relatively large hidden layers in their scoring functions (the s2e model has 26 layers and 494M parameters). Thus, one avenue for improving coreference speed is by reducing the model size. Additionally, while batching computations can improve parallelism and thus also throughput, the implementation of batching long documents of varying lengths is often sub-optimal, and results in many padded tokens which translate to wasted computation.

\subsection{Accurate Neural Coreference} Our recent \textsc{LingMess} model \citep{Otmazgin2022LingMessLI} improves coreference accuracy by observing that different types of entities require different strategies to score, and replacing the single mention-pair scorer with a set of specialized scorers. During inference, each mention pair is deterministically routed to one of the scorers, based on the the type of mentions being scored. This results in state-of-the-art coreference accuracy, while being somewhat less efficient to run and to batch than the \emph{s2e} model.
\section{Method}
\label{sec:method}

We employ two complementary directions in order to obtain a fast and efficient coreference model. First, we substantially reduce the size of the \emph{s2e} model using knowledge distillation~(§\ref{subsec:distillation}) from the \textsc{LingMess} model. Second, our implementation aims to maximize parallelism via batching while limiting the number of unnecessary computations such as padded tokens~(§\ref{subsec:efficient}). 


\subsection{Knowledge Distillation}
\label{subsec:distillation}

Knowledge Distillation is the process of learning a small student model from a large teacher model. 
\paragraph{Teacher model} We use the state-of-the-art \textsc{LingMess} model of \citet{Otmazgin2022LingMessLI} as the teacher model.
\paragraph{Student model} we build our student model as a variant of the \emph{s2e} model with fewer layers and parameters. The ``expensive'' Longformer \cite{Beltagy2020LongformerTL} encoder was replaced with DistilRoBERTa \cite{Sanh2019DistilBERTAD}, which is on average $\times 8$ faster than Longformer. The number of parameters of the mention and the antecedent pairwise scorers was reduced by a factor of 6. This reduces the total number of parameters from 494M to 91M. In addition, the number of sequential layers in the network reduced from 26 layers to only 8 layers (6 encoder layers, 1 mention scorer and 1 antecedent scorer). As a result, our student combines the strengths of the \emph{s2e} model by not constructing span representation with a lightweight encoder and substantially less model parameters. 
 \paragraph{Hard distillation}
Traditional approaches for knowledge distillation trains the student on the logits of the teacher model's predictions on unlabeled data \cite{Gou2021KnowledgeDA}. However, as we will further elaborate in Section~§\ref{subpar:soft_hard}, applying such an approach to a coreference model with all its components (i.e. encoder, mention scorer, pruning, antecedent scorer) achieves poor performance. To remedy this issue, we employ \textit{hard} target knowledge distillation, where the teacher model acts as an annotator for the unlabeled data and the student model learns from these ``silver'' annotations.

\begin{figure*}
    \centering
    \includegraphics[width=\textwidth]{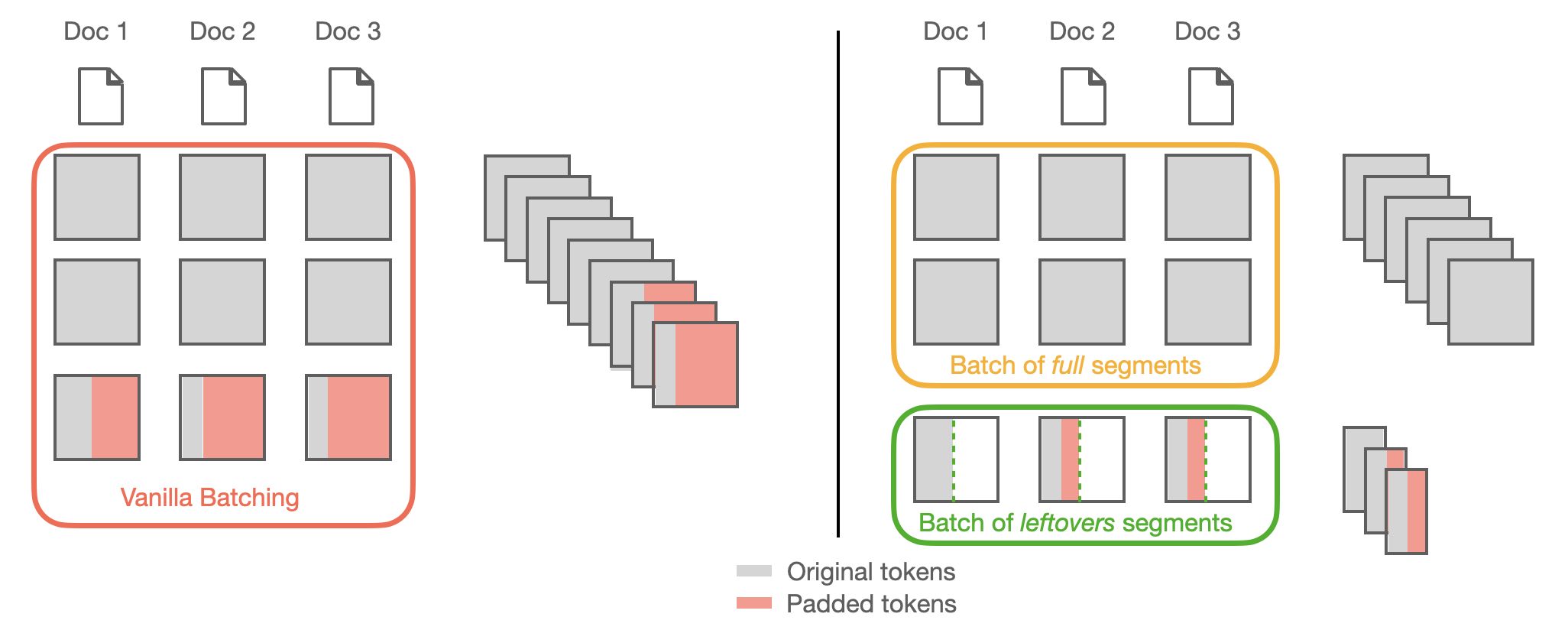}
    \caption{Illustration of document batching using vanilla batching (left) and our leftovers approach (right). In our leftover batching, we create two separate batches, one for the full segments without padding (orange) and one for the leftover segments (green), thus substantially reducing the number of padded tokens (red) compared to the vanilla approach.}
    \label{fig:leftover}
\end{figure*}

\subsection{Maximizing Parallelism and Reducing Unnecessary Computations}
\label{subsec:efficient}

\paragraph{Mention pruning}

As mentioned in Section~\ref{sec:bg}, the coreference model computes antecedent scores only for the $\lambda T$ spans with highest mention scores, where $T$ is the number of tokens. As the number of coreferring spans cannot be known in advance, the common approach is to use a soft pruning coefficient ($\lambda=0.4$) to guarantee high mention recall, while expecting the antecedent scorer to assign a negative score to pairs involving a wrong mention span~\citep{lee-etal-2017-end, kirstain-etal-2021-coreference}. In \model{}, we adopt a more aggressive pruning ($\lambda = 0.25$), which decreases the number of pairwise comparisons by a factor of 2.56 without harming performance.

\paragraph{Dynamic batching}
We adopt a dynamic batching approach which, given a large number of documents, batches documents until we reach a certain maximum number of tokens. Compared with the naive approach of batching a fixed number of documents together, dynamic batching enables to fully exploit the available memory in our hardware (this approach was also used  by \citet{kirstain-etal-2021-coreference} when \emph{training} the \emph{s2e} model, but not for inference).

\paragraph{Leftovers batching}
Figure~\ref{fig:leftover} illustrates our document batching strategy, in comparison with the common approach.

As mentioned in Section~\ref{sec:bg}, the first step in our coreference model consists of encoding the document using a transformer-based encoder. The common approach for encoding long documents with transformer encoders is to split the document into non-overlapping segments of \textit{max\_length}, where each segment is encoded separately \cite{joshi-etal-2019-bert, xu-choi-2020-revealing}. With that approach, for each long document, we obtain two\footnote{A document with fewer tokens than \textit{max\_length} has only one segment.} types of segment lengths: (1) one or more segments of \textit{max\_length} i.e. \textsc{full} tokens segment, (2) exact one segment $\leq$ \textit{max\_length}, i.e the \textsc{leftovers} tokens segment.

Then to batch multiple documents, a naive but popular approach consists of padding each document's \textsc{leftovers} segment to \textit{max\_length}. This results in a high number of padded tokens, e.g., 34.7\% of all tokens are padded when batching 2802 OntoNotes \cite{pradhan-etal-2012-conll} training set documents with \textit{max\_length} $=512$. Padded tokens result in unnecessary computations in each layer of the network, as well as unnecessary memory allocation.

To avoid such unnecessary computations, we split each batch into two batches such that the first batch is for the \textsc{full} segments and the second batch is for the \textsc{leftovers} tokens segment of each document. Then we pad the second batch to the max leftovers length rather than padding the leftovers segments to \textit{max\_length}. Finally we run the two batches separately and combine them afterwards. With this technique, the padded tokens in the OntoNotes training set reduced dramatically to 0.6\%. It should be noted that the aforementioned batching technique is not specific for coreference resolution and can be applied for other tasks that require processing long documents.
\begin{table*}[!t]
\small
\centering
\resizebox{\textwidth}{!}{
    \begin{tabular}{@{}lcccccccccccccc@{}}
    \toprule
    \multicolumn{1}{c}{} & & \multicolumn{3}{c}{MUC} & & \multicolumn{3}{c}{B\textsuperscript{3}} & & \multicolumn{3}{c}{CEAF\textsubscript{$\phi_4$}}\\
     \cmidrule{3-5} \cmidrule{7-9} \cmidrule{11-13}
       & & P & R & F1 & & P & R & F1 & & P & R & F1 & & Avg. F1 \\
    \midrule
    
    \citet{joshi-etal-2020-spanbert}        & & 85.8 & 84.8 & 85.3 & & 78.3 & 77.9 & 78.1 & & 76.4 & 74.2 & 75.3 & & 79.6 \\
    \citet{kirstain-etal-2021-coreference}  & & 86.5 & 85.1 & 85.8 & & 80.3 & 77.9 & 79.1 & & 76.8 & 75.4 & 76.1 & & 80.3 \\
    \citet{dobrovolskii-2021-word} &  & 84.9 & 87.9 & 86.3 & & 77.4 & 82.6 & 79.9 & & 76.1 & 77.1 & 76.6 & & 81.0 \\
    \citet{Otmazgin2022LingMessLI} (Teacher) &  & 88.1 & 85.1 & 86.6 & & 82.7 & 78.3 & 80.5 & & 78.5 & 76.0 & 77.3 & & 81.4 \\
    \midrule
    \model{} OntoNotes only  & & 78.5 & 84.3 & 81.3 & & 68.2 & 74.8 & 71.4 & & 64.1 & 72.9 & 68.2 & & 73.7 \\
    \model{} Multi-News      & & 84.8 & 82.8 & 83.4 & & 76.8 & 73.7 & 75.2 & & 73.8 & 72.7 & 73.2 & & 77.4 \\
    \quad + FT OntoNotes     & & 85.0 & 83.9 & 84.4 & & 77.6 & 75.5 & 76.6 & & 74.7 & 74.3 & 74.5 & & 78.5 \\
    \bottomrule
    \end{tabular}
}
\caption{Performance on the test set of the English OntoNotes 5.0 dataset. The averaged F1 of MUC, B\textsuperscript{3}, CEAF\textsubscript{$\phi$} is the main evaluation metric. }
\label{table:results}
\end{table*}
\begin{table}[t!]
\small
\centering
\resizebox{0.49\textwidth}{!}{
\begin{tabular}{@{}lcccc@{}}
\toprule
   & \textbf{Masc} & \textbf{Fem} & \textbf{Bias} & \textbf{Overall} \\
\midrule

\citet{Otmazgin2022LingMessLI} (Teacher)    & 91.3 & 87.8 & 0.96 & 89.6  \\
\model{}                                    & 87.8 & 83.5 & 0.95 & 85.7  \\
\bottomrule
\end{tabular}}
\caption{Performance on the test set of the GAP coreference dataset. The reported metrics are F1 scores.}
\label{table:gap}
\end{table}

\begin{table}[t]
\begin{threeparttable}[b]
\small
\centering
\resizebox{0.48\textwidth}{!}{

\begin{tabular}{@{}lcc@{}}
\toprule

\textbf{} & \textbf{Runtime} & \textbf{Memory} \\
\midrule



\citet{joshi-etal-2020-spanbert}\tnote{1}                   & 12:06 & 27.4 \\
\citet{Otmazgin2022LingMessLI} (Teacher)                    & 06:43 & 4.6 \\ \quad + Batching\tnote{2}                                           & 06:00 & 6.6 \\

\citet{kirstain-etal-2021-coreference}                      & 04:37 & 4.4 \\
\citet{dobrovolskii-2021-word}                              & 03:49 & 3.5 \\
\model{}                                                    & 00:45 & \textbf{3.3} \\
\quad + Batching\tnote{2}                                   & 00:35 & 4.5 \\
\quad \quad + Leftovers batching\tnote{2}                   & \textbf{00:25} & 4.0 \\

\bottomrule
\end{tabular}}
\begin{tablenotes}
       \item [1] AllenNLP package implementation.
       \item [2] 10K tokens in a single batch.
\end{tablenotes}
\caption{The inference time(Min:Sec) and memory(GiB) for each model on 2.8K documents. Average of 3 runs. Hardware, NVIDIA Tesla V100 SXM2.} 

\label{table:efficiency}
\end{threeparttable}
\end{table}


\begin{table}[t!]
\small
\centering
\resizebox{0.42\textwidth}{!}{
\begin{tabular}{@{}lcccc@{}}
\toprule
   & \textbf{\#Docs} & \textbf{\#Chains} & \textbf{\#Mentions} \\
\midrule


OntoNotes           & 2.8K & 35K & 155K  \\
Multi-News          & 123K & 2M & 9M  \\
\bottomrule
\end{tabular}}
\caption{Coreference statistics of the training set of OntoNotes and Multi-News.}

\label{table:ds_comp}
\end{table}

\section{Experiments and Results}
\label{sec:results}

\paragraph{Experiments setup} In our experiments, we use the Multi-news \cite{fabbri-etal-2019-multi} dataset to train our teacher-student architecture. The Multi-news dataset is an open source dataset aimed at NLP summarization. Each entry in the dataset contains multiple documents and a summary of these documents. For our purposes, we ignored the summaries, and train our student model on the documents, a total of 123,227 documents in the news domain. We chose Multi-News because it contains a large number of documents in the news genre, which would result in a large number of coreference clusters (see Table~\ref{table:ds_comp} for statistics). Furthermore, we use the English portion of the OntoNotes \cite{pradhan-etal-2012-conll} dataset to evaluate the student model performance (on the test set) and to further fine-tune the student model (on the train set). 

The student training procedure includes three phases. In the first phase, we predict coreference clusters on MultiNews using the teacher model~\citep{Otmazgin2022LingMessLI}. Secondly, following~\citep{wu-etal-2020-corefqa, dobrovolskii-2021-word}, we pre-train the mention scorer of the student on the output mentions of the teacher, then train the full student model on the predicted teacher coreference clusters. Finally, we finetune the student model on the OntoNotes training set. 

\paragraph{Accuracy} Table~\ref{table:results} shows \model{}'s performance on the OntoNotes test set according to the standard evaluation metrics for coreference resolution. \model{} achieved 78.5 F1 with knowledge distillation and finetuning on OntoNotes. When trained only on OntoNotes, \model{} achieved only 73.7 F1 ($-$4.8 F1), showing a substantial benefit from knowledge distillation on the Multi-news dataset \cite{fabbri-etal-2019-multi}. In comparison with other coreference models, \model{} degrades by 1.1 point compared to the \citet{joshi-etal-2019-bert} model in the AllenNLP package, by 2.9 F1 points compared to \textsc{LingMess}~\citep{Otmazgin2022LingMessLI}, the teacher, and by 1.8 F1 points versus \citet{kirstain-etal-2021-coreference}, the \emph{s2e} model, which \model{} is a variant of. Table~\ref{table:gap} shows similar trends for the GAP dataset~\citep{webster-etal-2018-mind}. This degradation comes in favor to the model efficiency, which we will discuss in the next paragraph.

\paragraph{Speed and Memory Usage} 
Table~\ref{table:efficiency} summarizes the efficiency of the different coreference models. Using the techniques described in Section~§\ref{sec:method} which reduces the model size, maximize batching and avoids unnecessary computation, the inference time on 2.8K documents is significantly reduced by factor of 9 from the 03:49 minutes of \citet{dobrovolskii-2021-word}---the fastest model to date---to only 25 seconds for \model{}.\footnote{These experiments used batch sizes of 10k tokens. Increasing the batch sizes increase memory consumption, but does not improve overall speed on our NVIDIA Tesla V100 hardware.}

Most of the speed increase (80\%) is due to the smaller model size achieved through distillation, which alone reduces the runtime to 45 seconds. However, introducing batching further reduces the runtime by additional 22\% to 35 seconds, and our novel leftover batching reduces further 29\%, and gets us to 25 seconds. The more aggressive mention pruning (not shown in the table) had only a negligible additive effect in our experiments, reducing the runtime from 25 to 24 seconds. 
  
Without batching, \model{} also consumes less memory then \citet{dobrovolskii-2021-word}, a model that recently reduces the coreference complexity from $\mathcal{O}(n^4)$ to $\mathcal{O}(n^2)$.

Finally, compared to one of the most widely used coreference models, the  \citet{joshi-etal-2020-spanbert} model available through the AllenNLP package \cite{Gardner2017AllenNLP}, \model{} is 29 times faster and consumes 85\% less memory.

\subsection{Further Analysis}
\begin{figure}[t]
    \centering
    \includegraphics[scale=0.54]{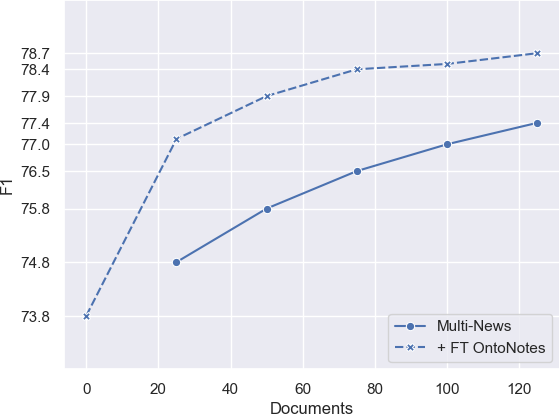}
    \caption{The knowledge distillation learning curve. The x-axis is the number of documents we took from the Multi-News dataset to train the student model. The y-axis is the student F1 score for each size.}
    \label{fig:kd_lr}
\end{figure}


\paragraph{Effect of Unlabeled Data Size}
We first analyze the effect of the amount of unlabeled data used in distillation, by training the student model on different amounts of training data. As Figure~\ref{fig:kd_lr} shows the performance gain between 25K documents and 50K documents is 1 F1 point while the gain between 100K and 125K documents decreases to 0.4 points. This indicates that the gain margin is decreasing, but overall, increasing the dataset size continuously improves the model performance and there is a room for improvement with more data. Fine-tuning the distilled model on the in-domain OntoNotes data consistently improves the results, but is also additive with the distillation: the fine-tuned performance also increases with more unlabaled data in distillation.

\paragraph{Effect of the Teacher Model}
To estimate the effect of the teacher model, we compare the \textsc{LingMess} teacher to a \emph{s2e} model~\citep{kirstain-etal-2021-coreference} teacher on the same unlabeled data. 
We obtain 76.6 F1 with \emph{s2e} (vs. 77.4 F1 with \textsc{LingMess}) on the OntoNotes (test set) when training only on Multi-News and 78.3 F1 with \emph{s2e} (vs. 78.5 F1 with \textsc{LingMess}) after further fine-tuning on OntoNotes (training set). This indicates that the student accuracy increases when more accurate models are used in knowledge distillation, even with hard labels.



\paragraph{Soft VS. Hard Distillation}
\label{subpar:soft_hard}

Our first attempt to transfer the coreference knowledge from the teacher model to the student model was the traditional knowledge distillation, i.e. soft targets knowledge distillation. For each example in the training set, we forward it first in the teacher model and obtained the top-scoring spans indices, and the pairwise coreference logits. Then we forward the example in the student network (at pruning stage we use the teacher's top-scoring spans indices) to obtain the student coreference pairwise logits. Following common training objective~\citep{Hinton2015DistillingTK, Sanh2019DistilBERTAD, jiao-etal-2020-tinybert}, we optimize the student model using the soft cross entropy loss between the student and the teacher logits.


Our student model reached only 64 F1, while achieving 73.7 F1 without knowledge distillation. Soft distillation presents challenges in coreference models. The main challenge we encounter is at the pruning stage, where both teacher and student should prune the exact same mentions from their individual mention scorer. This forces the student model to learn from a conditional antecedent distribution of the teacher spans indices instead of the full antecedent distribution. 

Additionally, we observe that learning to mimic the teacher logits may trouble the training because logits can violate transitivity (e.g positive score for the mention pairs $(a, b)$ and $(b, c)$ but a negative score for $(a, c)$) and propagate contradictory information. Specifically, we build the coreference clusters based on the positive pairwise scores, and verify whether all pairwise scores within the same coreference clusters are positive, as we would naturally expect. In fact, 53.8\% of the pairwise scores within the same coreference cluster are negative. In contrast, this undesired behavior does not happen in hard distillation because we assign positive labels for all coreferring antecedents for each mention.





\section{Conclusions}
We introduce the \emph{fastcoref} python package for coreference resolution, and hope its speed an ease of use will facilitate work that utilizes coreference resolution at scale.

\section*{Acknowledgements}
This project has received funding from the European Research Council (ERC) under the European
Union’s Horizon 2020 research and innovation
programme, grant agreement No. 802774 (iEXTRACT). Arie Cattan is partially supported by the Data Science Institute at Bar-Ilan University.

\bibliographystyle {acl_natbib}
\bibliography{anthology,custom}

\end{document}